\documentclass[10pt,onecolumn]{IEEEtran}
\usepackage{cite}
 \usepackage[pdftex]{graphicx}
  \graphicspath{{../pdf/}{../jpeg/}}
   \DeclareGraphicsExtensions{.pdf,.jpeg,.png}

\usepackage{amsmath}
\usepackage{amssymb}
\usepackage{bm}
\usepackage{bbm}
\usepackage{array}

\ifCLASSOPTIONcompsoc
  \usepackage[caption=false,font=normalsize,labelfont=sf,textfont=sf]{subfig}
\else
  \usepackage[caption=false,font=footnotesize]{subfig}
\fi

\usepackage{url}

\usepackage[shortcuts,acronym]{glossaries}

\usepackage{multirow}

\usepackage{tabularx}
\usepackage{booktabs}

\usepackage{siunitx}
\usepackage{lineno,hyperref}
\usepackage{colortbl}
\usepackage[dvipsnames]{xcolor}
\hypersetup{
    colorlinks=true,
    linkcolor = ForestGreen,
    citecolor = RoyalBlue,
    anchorcolor = cyan,
}

\usepackage{textcomp}

\makeglossaries
\newacronym{1d}{2D}{one-dimensional}
\newacronym{2d}{2D}{two-dimensional}
\newacronym{3d}{3D}{three-dimensional}
\newacronym{sc}{S/C}{spacecraft}

\newacronym{adr}{ADR}{Active Debris Removal}
\newacronym{ai}{AI}{Artificial Intelligence}
\newacronym{ann}{ANN}{Artificial Neural Network}
\newacronym{alhat}{ALHAT}{Autonomous Landing Hazard Avoidance Technology}
\newacronym{ap}{AP}{Average Precision}

\newacronym{bce}{BCE}{Binary Cross-Entropy}
\newacronym{bn}{BN}{batch normalization}

\newacronym{cbam}{CBAM}{Convolutional Block Attention Module}
\newacronym{cl}{CL}{Convolutional Layer}
\newacronym{cnn}{CNN}{Convolutional Neural Network}
\newacronym{coco}{COCO}{Common Objects in Context}
\newacronym{ciou}{CIOU}{Complete Intersection over Union}
\newacronym{cd}{CD}{Crater Detection}
\newacronym{ci}{CI}{Crater Identification}
\newacronym{cro}{CRO}{Candidate for a Regional Object}

\newacronym{deeptam}{DeepTAM}{Deep Tracking and Mapping}
\newacronym{dmlp}{DMLP}{Deep Multi-Layer Perception}
\newacronym{dem}{DEM}{Digital Elevation Map}
\newacronym{dtm}{DTM}{digital terrain model}
\newacronym{dl}{DL}{Deep Learning}
\newacronym{drcnn}{DRCNN}{Deep Recurrent Convolutional Neural Network}
\newacronym{dnn}{DNN}{Deep Neural Network}

\newacronym{iou}{IOU}{Intersection over Union}
\newacronym[longplural={Degrees-of-Freedom}]{dof}{DoF}{Degree-of-Freedom}
\newacronym{ilrs}{ILRS}{International Lunar Research Station}

\newacronym{ekf}{EKF}{Extended Kalman Filter}
\newacronym{esa}{ESA}{European Space Agency}

\newacronym{fl}{FCL}{Fully Connected Layer}
\newacronym{fpr}{FPR}{False Positive Rate}
\newacronym{fpga}{FPGA}{Field-Programmable Gate Array}

\newacronym{gmm}{GMM}{Gaussian Mixture Modelling}
\newacronym{gnc}{GNC}{Guidance, Navigation and Control}
\newacronym{gpops}{GPOPS II}{General Purpose Optimal Control Software}
\newacronym{gpu}{GPU}{Graphics Processing Unit}

\newacronym{prc}{PRC}{precision-recall curve}

\newacronym{hda}{HDA}{Hazard Detection and Avoidance}
\newacronym{hrnet}{HRNet}{High-Resolution Net}

\newacronym{icp}{ICP}{Iterative Closest Point}
\newacronym{ilsvrc}{ILSVRC}{ImageNet Large Scale Visual Recognition Challenge}

\newacronym{krn}{KRN}{Keypoint Regression Network}
\newacronym{kpec}{KPEC}{Kelvins Pose Estimation Challenge}

\newacronym{lclf}{LCLF}{Lunar-Centred, Lunar-Fixed Coordinates}
\newacronym{log}{LoG}{Laplacian of Gaussian}
\newacronym{lro}{LRO}{Lunar Reconnaissance Orbiter}
\newacronym{lstm}{LSTM}{Long Short-Term Memory}
\newacronym{lvlh}{LVLH}{Local-Vertical, Local-Horizontal}

\newacronym{mnist}{MNIST}{Modified National Institute of Standards and Technology}
\newacronym{ml}{ML}{Machine Learning}
\newacronym{mlp}{MLP}{Multilayer Perceptron}
\newacronym{mse}{MSE}{Mean Square Error}
\newacronym{mcmf}{MCMF}{Moon-Centered Mood-Fixed}

\newacronym{nasa}{NASA}{National Aeronautics and Space Administration}
\newacronym{nn}{NN}{Neural Network}
\newacronym{nea}{NEA}{Near-Earth Asteroid}
\newacronym{nst}{NST}{Neural Style Transfer}

\newacronym{odn}{ODN}{Object Detection Network}

\newacronym{pcc}{PCC}{Pearson correlation coefficient}
\newacronym{pds}{PDS}{Planetary Data System}
\newacronym{ppv}{PPV}{Positive Predictive Value}
\newacronym{pycda}{PyCDA}{Python Crater Detection Algorithm}
\newacronym{pnp}{P$n$P}{Perspective-n-Point}
\newacronym{pilot}{PILOT}{Precise Intelligent Landing using On-board Technology}

\newacronym{ransac}{RANSAC}{Random Sample Consensus}
\newacronym{rcnn0}{R-CNN}{Region-based Convolutional Neural Network}
\newacronym{rcnn}{RCNN}{Recurrent Convolutional Neural Network}
\newacronym{rgb}{RGB}{Red-Green-Blue}
\newacronym{roi}{RoI}{Region of Interest}
\newacronym{rpn}{RPN}{Region Proposal Network}
\newacronym{rmse}{RMSE}{Root Mean Square Error}
\newacronym{rnn}{RNN}{Recurrent Neural Network}
\newacronym{rfcn}{R-FCN}{region-based fully convolutional networK}

\newacronym{senet}{SENet}{Squeeze-and-Excitation Network}
\newacronym{sift}{SIFT}{Scale Invariant Feature Transform}
\newacronym{slam}{SLAM}{Simultaneous Localisation and Mapping}
\newacronym{sgd}{SGD}{Stochastic Gradient Descent}
\newacronym{speed}{SPEED}{Spacecraft Pose Estimation Dataset}
\newacronym{spn}{SPN}{Spacecraft Pose Network}
\newacronym{ssd6d}{SSD-6D}{Single Shot Detector 6D}
\newacronym{soc}{SoC}{System-on-a-Chip}

\newacronym{tpr}{TPR}{True Positive Rate}
\newacronym{trn}{TRN}{Terrain Relative Navigation}
\newacronym{tir}{TIR}{thermal infrared}

\newacronym{urso}{URSO}{Unreal Rendered Spacecraft On-Orbit}

\newacronym{vo}{VO}{Visual Odometry}

\newacronym{wac}{WAC}{Wide Angle Camera}
\newacronym{xai}{XAI}{explainable artificial intelligence}

\usepackage{multirow}
\begin{document}
%
\title{Explainable Convolutional Networks for Crater Detection and Lunar Landing Navigation}

%
%
%

\author{Jianing~Song,
        Nabil~Aouf,
        Duarte~Rondao,
        Christophe~Honvault,
        and~Luis~Mansilla
\thanks{J. Song is a Postdoctoral Research Fellow with the Department of Electrical and Electronic Engineering at City, University of London, ECV1 0HB London, UK (e-mail: jianing.song@city.ac.uk).}
\thanks{N. Aouf is a Professor of Robotics and Autonomous Systems with the Department of Electrical and Electronic Engineering at City, University of London, EC1V 0HB, UK (nabil.aouf@city.ac.uk).}
\thanks{D. Rondao is a Postdoctoral Research Fellow with the Department of Electrical and Electronic Engineering at City, University of London, EC1V 0HB, UK (e-mail: duarte.rondao@city.ac.uk).}
\thanks{L. Mansilla is a Developer at the European Space Research and Technology Centre (ESTEC), ESA, Postbus 299. 2200 AG Noordwijk, The Netherlands (e-mail: luis.mansilla@ext.esa.int)}
\thanks{C. Honvault is the head of software technology section at ESA, Keplerlaan 1, 2201AZ Noordwijk, The Netherlands (e-mail: christophe.honvault@esa.int).}
}


\maketitle

\begin{abstract}
The Lunar landing has drawn great interest in lunar exploration in recent years, and autonomous lunar landing navigation is fundamental to this task. \gls{ai} is expected to play a critical role in autonomous and intelligent space missions, yet human experts question the reliability of \gls{ai} solutions.Thus, the \gls{xai} for vision-based lunar landing is studied in this paper, aiming at providing transparent and understandable predictions for intelligent lunar landing. Attention-based Darknet53 is proposed as the feature extraction structure. For crater detection and navigation tasks, attention-based YOLOv3 and attention-Darknet53-LSTM are presented respectively. The experimental results show that the offered networks provide competitive performance on relative crater detection and pose estimation during the lunar landing. The explainability of the provided networks is achieved by introducing an attention mechanism
into the network during model building. Moreover, the \gls{pcc} is utilised to quantitively evaluate the explainability of the proposed networks, with the findings showing the functions of various convolutional layers in the network.

\end{abstract}


%

\section{Introduction}
Exploring and landing on the Moon has long been a challenge of great interest in space science and technology. \gls{esa} has developed the Lunar exploration programme based on four main missions, with Luna Resurs, Orion, ISRU, and Heracles, in which innovative European technology will be introduced to these missions to achieve success. \gls{nasa} has launched the Artemis mission, aiming at using innovative technologies to explore more for scientific discovery, economic benefits, and inspiration for a new generation of explorers. China landed the Chang'e-4 lunar probe on the Moon in 2019 and will continue the lunar exploration program to build the \gls{ilrs} with intelligent techniques.

To achieve future lunar missions, the ability to autonomous landing is required, and autonomous landing navigation is fundamental to this task. However, no autonomous precision landing on the Moon has been performed\cite{maass2020crater}.  Apollo landing navigation was achieved by manually locating and matching craters by the pilots. 
Later, craters were considered to be ideal natural markers that can be used for navigation purposes. For example, the Mars 2020 mission of \gls{nasa} is going to match landmarks (maybe craters) from onboard images to reduce the landing error to less than \SI{40}{\meter} \cite{maass2020crater,johnson2017lander}. In the Luna Resurs mission conducted by the Russian agency Roscosmos, India, and \gls{esa}, the precise and safe landing on the Moon named \gls{pilot} will be offered by \gls{esa}. The \gls{pilot} is a state-of-the-art system developed for the \gls{esa} Lunar Lander project, which represents the first European initiative aimed at achieving a precision landing via crater matching \cite{hamel2018validation}. Additionally, autonomous optical navigation is gaining great attention in future lunar and planetary landings. 

Meanwhile, \gls{ai} has achieved great success in various areas and is becoming increasingly attractive to the space research community. Intelligent algorithms, through the recent developments of \gls{ai}, are also making enormous progress in providing intelligence to space missions. These \gls{ai}-based algorithms attain promising results in various scenarios, from rendezvous and lunar landing to asteroid exploration. A recent review of \gls{dl}-based spacecraft relative navigation has shown considerable \gls{dl}-based crater detection and landing navigation methods \cite{song2022deep}.
However, the trust in freely deploying them on space vehicles to increase their autonomy is questionable by the most qualified scientists in the space domain.
A thorough search of the relevant literature yielded a few related articles that covered explainability. Therefore, the objective of this paper is to study \gls{xai}, with a special focus on explainable methods \gls{dl} in the estimation of relative poses based on vision, providing transparent and understandable predictions for intelligent lunar landing navigation.

In this paper, an explainable end-to-end framework is presented to provide pose estimates from a sequence \gls{rgb} image during the Lunar landing. We first study the attention-based crater detection network taking advantage of YOLOv3 and the attention mechanism. Next, the backbone of the trained crater detection network is extracted and applied with \gls{lstm} to a plausible landing navigation network. The explainability of the network is achieved by the attention layers, which show how the network focuses on the features related to the given label. Moreover, the Pearson correlation coefficient is utilised to quantitively evaluate the explainability of the proposed network.

The rest of the paper is organised as follows: Section \ref{sec:relatedwork} is devoted to introducing the previous work and techniques related to the paper. Section \ref{sec:methodlogy} details the proposed neural network pipeline and network training strategies. Section \ref{sec:scenario definition} describes the studied scenario and simulation tools. Section \ref{sec:experiments} offers experimental results from the synthetic images and discussion. Section \ref{sec:conclusion} outlines our conclusions and future work directions.

\section{Related work}
\label{sec:relatedwork}
Craters are ideal landmarks for autonomous navigation on the Moon, and the development of optical techniques makes crater-based autonomous lunar landing a popular and reliable method. With recent developments in \gls{dl}, there has been renewed interest in using \gls{dl}-based crater detection for landing navigation.

\subsection{DL-based crater detection}
For the purpose of intelligent crater detection, earlier studies used \gls{cnn} as a classifier to validate selected features \cite{emami2015automatic, cohen2016crater, palafox2017automated}. An open-source crater detection library \gls{pycda} employed a downsized U-Net framework to detect new craters that had never been catalogued \cite{klear2018pycda}. Wang et al. proposed an end-to-end full \gls{cnn}, CrateIDNet, for simultaneous crater detection and identiﬁcation \cite{wang2018crateridnet}. Silburt et al. presented DeepMoon (U-shaped) architecture that worked on \gls{dem} for robust crater detection on the lunar surface \cite{silburt2019lunar}. Taking advantage of \gls{dem} images, \cite{Moon2021Jia} developed an attention-aware U-Net for Lunar crater detection, \cite{ELCD2022Fan} offered an attention-based encoder-decoder framework for Lunar crater detection. Additionally, \cite{Mars2022Li} utilised YOLOv5 to identify craters from Mars \gls{tir} images. \cite{yang2020lunar} studied a standard \gls{rfcn} for crater detection from three-channel images generated by optical images, slope maps, and profile curvature. 
Similarly, \cite{Automated2022Atal} combined the optical images, \glspl{dem}, and slope maps and then passed them through a standard Mask \gls{rcnn0} for crater detection.  
\cite{latorre2023transfer} provided a real-time crater detention using the U-Net from Ceres \gls{dtm} by taking advantage of transfer learning from the Moon to Ceres. \cite{zhang2023planet} offered an automated pipeline named CraterNet to detect craters on the Moon and Mercury from \gls{dem} images in an unsupervised domain adaption fashion.


\subsection{DL-based lunar landing navigation}

For the purpose of \gls{dl}-based Lunar landing navigation, 
Downes et al. proposed a LunaNet framework to detect craters from greyscale images \cite{downes2020}, which is the same as the DeepMoon framework. 
In 2020, Downes et al. applied the LunaNet in a terrain navigation system to robustly estimate relative position and velocity information \cite{Lena8}. Later, \cite{Crater2021Chen} proposed a dense point crater detection network that matches the detected craters to a database and estimated relative poses using \gls{pnp} for both orbit around and landing scenarios. 
\cite{silvestrini2022optical} used an \gls{odn} to extract observed craters, followed by image processing algorithms for pseudo-measurements. The absolute position is estimated by an extended Kalman filter, with the estimated horizontal and vertical errors of around \SI{200}{\meter} along the reference trajectory. Few published studies provide lunar landing navigation solutions in an end-to-end manner. The only work available in \cite{chekakta2022robust} offered a hybrid \gls{drcnn} combining a \gls{cnn} and \gls{lstm} for autonomous space landers from \gls{3d} LiDAR data. In \cite{chekakta2022robust}, the proposed \gls{cnn} network was composed of three convolutional layers aimed at extracting sequences of feature vectors from input LiDAR images, and the \gls{lstm} was intended to model the spacecraft landing navigation automatically. Recent research has suggested the advantage of the end-to-end \gls{dl} framework in feature extraction and robust modelling of navigation kinematics.


\subsection{Attention mechanism}

The attention mechanism was born to investigate the links between inputs and outputs of \gls{dl} models in the ﬁeld of natural language processing \cite{bahdanau2014neural}. Attention mechanisms were initially used in machine translation achieving notable success, and later expanded to computer vision with various types explored \cite{xu2015show}, \cite{vaswani2017attention}.

In \cite{jaderberg2015spatial}, a \gls{cnn}-\gls{lstm} encoder-decoder pipeline was used to extract image features and generate descriptive phrases, with the attention mechanism for weights learning. Visualising the attention weights revealed the model paying attention to which portions of the image produce a speciﬁc word. The research introduced the distinction between soft and hard attention, depending on whether the attention had access to the complete image or merely a patch of it. The alignment weights of soft attention are learned and softly assigned to the whole patches in the source image, whereas hard attention only attends to one patch of the image at a time.
The soft attention-based model is smooth and differentiable, but the hard attention-based model is non-differentiable and trained by reinforcement learning. Therefore, we focus on soft attention methods in this paper. 

\subsubsection{CBAM Attention}
\begin{figure}[htb]
    \centering
    \includegraphics[width=0.6\linewidth]{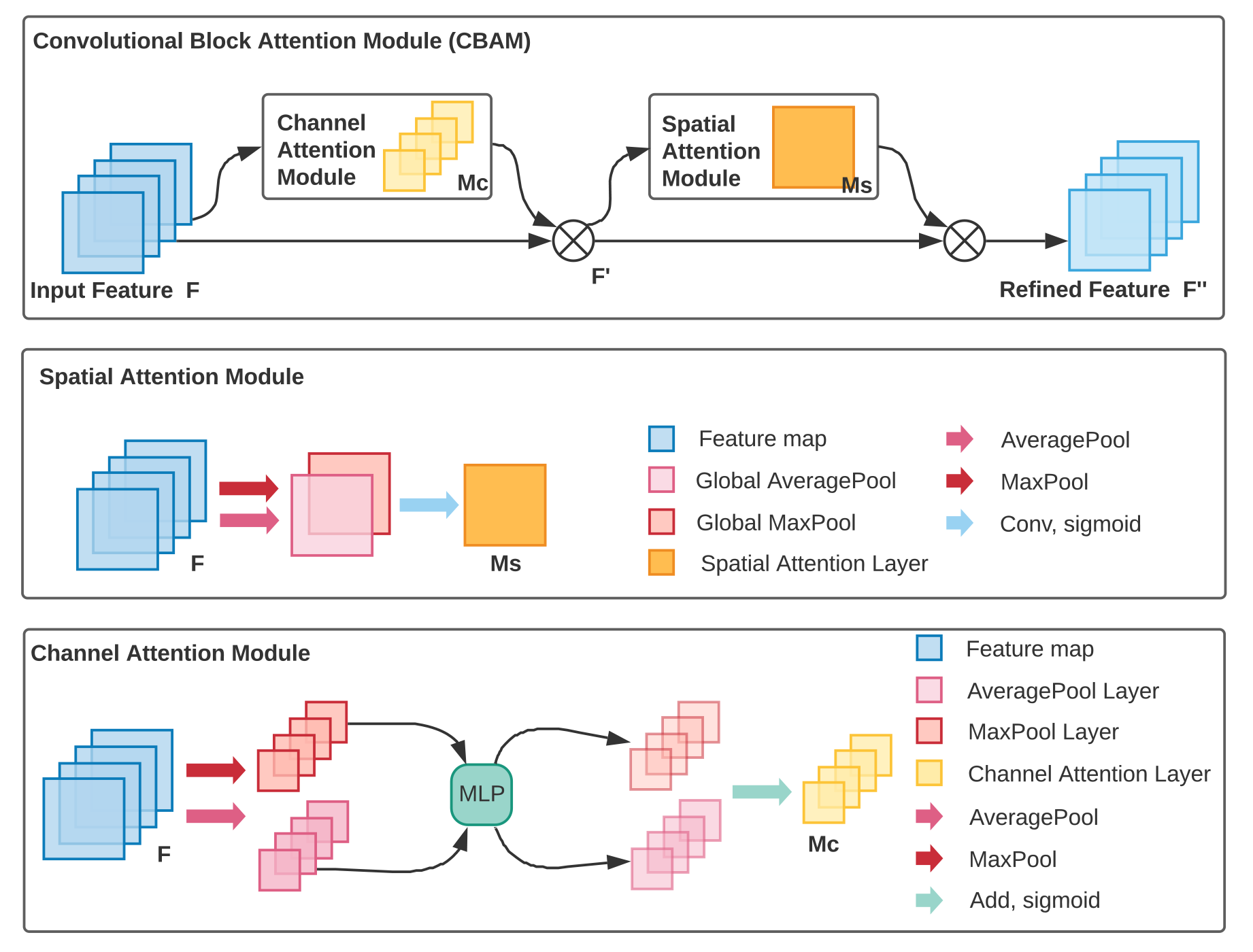}
    \caption{Attention module of \gls{cbam}}
    \label{fig:cbam}
\end{figure}

\cite{woo2018cbam} studies the spatial and channel attention structure named \gls{cbam}, showing in Fig. \ref{fig:cbam}. Assuming an intermediate feature map $\bm{F} \in C \times H \times W$ as input, \gls{cbam} sequentially infers a \gls{1d} channel attention map $\bm{M}_c \in C \times 1 \times 1$ and a \gls{2d} spatial attention map
$\bm{M}_s \in 1 \times H \times W$. The following is a summary of the overall attention process.
\begin{equation}
    \begin{split}
        \bm{F}' &= \bm{M}_c(\bm{F}) \otimes \bm{F}\\
        \bm{F}'' &= \bm{M}_s(\bm{F}') \otimes \bm{F}'
    \end{split}
\end{equation}
where $\otimes$ represents element-wise multiplication and $\bm{F}''$ is the refined feature output.

\subsubsection{Spatial Attention}
Spatial Attention \cite{jaderberg2015spatial} generates spatial attention in the convolutional neural networks by utilising the inter-spatial relationship of features. It focuses on where there is an informative part, which is complementary to channel attention. Generally, to compute the spatial attention of \gls{cbam}, there are steps: (1) Both global max-pooling and global average-pooling operations are applied to input features along the channel axis and (2) concatenated to generate an efﬁcient feature descriptor. (3) Applying convolution operation, the dimension of the channel decreases to 1. (4) The above results are fed to the Sigmoid function to generate spatial attention features. The spatial attention features are then multiplied with the input feature map to increase spatial attention. The mathematical expression of spatial attention is
\begin{equation}
\begin{split}
      \bm{M}_s(\bm{F}) &= \sigma ( f^{7\times7} ( [\text{AvgPool}(\bm{F}); \text{MaxPool}(\bm{F})]))  \\
      &= \sigma (f^{7\times7} ([\bm{F}_\text{avg}^s;\bm{F}_\text{max}^s]))
\end{split}
\end{equation}
where $\sigma$ denotes the Sigmoid function, $f^{7\times7}$ is a $7\times7$ kernel of a convolutional block, $\bm{F}_\text{avg}^s \in \mathbb{R} ^{1\times H \times W}$, $\bm{F}_\text{max}^s \in \mathbb{R} ^{1\times H \times W}$

\subsubsection{Channel Attention}
Channel attention is developed to boost the representational power of a network by enhancing spatial encoding. \cite{hu2018squeeze} investigates the channel relationship and proposes the \gls{senet}. The channel attention module of \gls{senet} consists of three parts: squeeze, excitation, and scaling(attention). 
The channel attention module in \cite{woo2018cbam} (see Fig. \ref{fig:cbam}) is nearly identical to \cite{hu2018squeeze}, except that MaxPool is added to get more unique channel characteristics. The methodology of channel attention in \cite{woo2018cbam} is: (1) Performing both average-pooling and max-pooling on the input features, generating two different spatial descriptors denoted as $\bm{F}_\text{avg}^c$ and $\bm{F}_\text{max}^c$, respectively. (2) Feeding both descriptors to a shared \gls{mlp}, where the number of channels is ﬁrst compressed and then reconstructed. (3) Element-wise summation of the output feature vectors from \gls{mlp}. The Sigmoid function is then connected to generate attention weights of 0~1 between channels. (4) Multiplying the scale back to the original input feature. Mathematically, the channel attention of \gls{cbam} is computed as,
\begin{equation}
\begin{split}
      \bm{M}_c(\bm{F}) &= \sigma ( \text{MLP} ( \text{AvgPool}(\bm{F})) + \text{MLP}( \text{MaxPool}(\bm{F})))  \\
      &= \sigma (\bm{W}_1 (\bm{W}_0 ( \bm{F}_\text{avg}^c)) + 
      \bm{W}_1 (\bm{W}_0 ( \bm{F}_\text{max}^c)))
\end{split}
\end{equation}
where \gls{mlp} weights are $\bm{W}_0 \in C/r \times C$ and $\bm{W}_1 \in C \times C/r $, followed by ReLU activation function.

\section{Methodology}
\label{sec:methodlogy}
This section details the proposed explainable crater detection and Lunar landing navigation networks. The attention-based YOLOv3 is used for crater detection, while the attention-based \gls{rcnn} network is designed to achieve end-to-end Lunar landing navigation. Both networks employ the same feature extraction framework noted as Attention-based Darknet53, where The incorporated attention mechanism is used to achieve explainability. 

\subsection{Attention-based CNN for feature extraction}
To learn features with explainability, a \gls{cnn} adaptive with attention mechanism is developed to perform explainable feature extraction on input monocular \gls{rgb} images. Among various powerful \glspl{cnn} used for feature extraction, such as ResNet, Inception-v3, and Faster \gls{rcnn0}, the Darknet53 is selected as the backbone of the explainable CNN to extract features efficiently. The Darknet53 comprises a series of convolution layers at dimensions of $1\times1$ and $3\times3$, with a total of 53 layers (including the last fully connected layer but excluding the residual layer). Each convolution layer follows a \gls{bn} layer and a LeakyReLU layer. The residual network modules in Darknet53 aim to solve the gradient disappearance or gradient explosion problems. Thus, we can more easily control the propagation of the gradient and perform network training.

\begin{figure}[htb]
    \centering
    \includegraphics[width=0.6\linewidth]{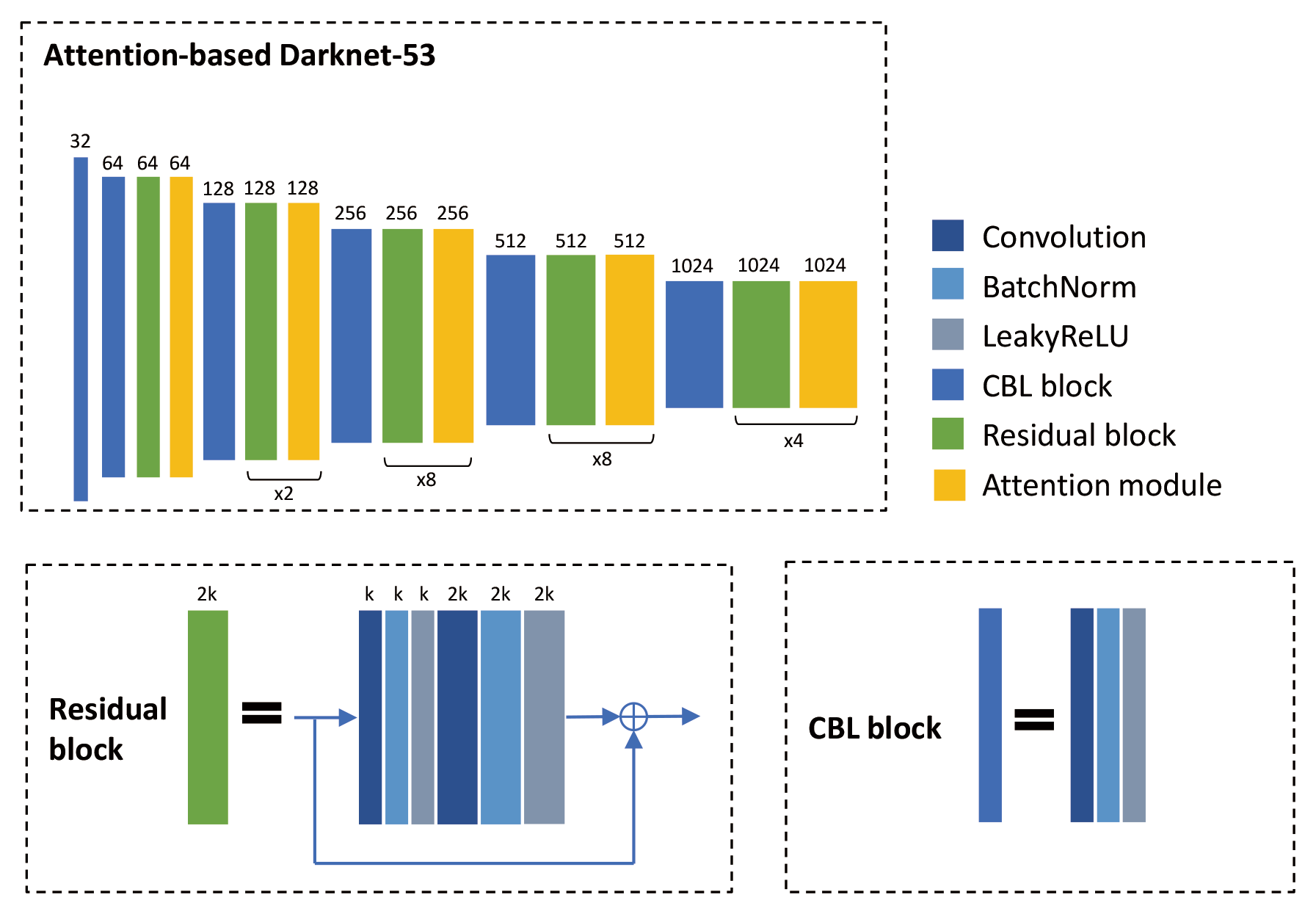}
    \caption{Proposed attention-based Darknet53 framework}
    \label{fig:att-dk53}
\end{figure}

Considering the crater detection and navigation tasks, the ﬁrst and last layers of Darknet53 are modiﬁed due to the varied size of input images and expected outputs. Further, the \gls{cbam} modules are added following each residual block to realise the explainability and understand how each block contributes to the predictions. Fig. \ref{fig:att-dk53} shows the proposed attention-based Darknet53 framework.

\subsection{Attention-based YOLOv3 for crater detection}
\begin{figure*}[htb]
    \centering
    \includegraphics[width=0.6\linewidth]{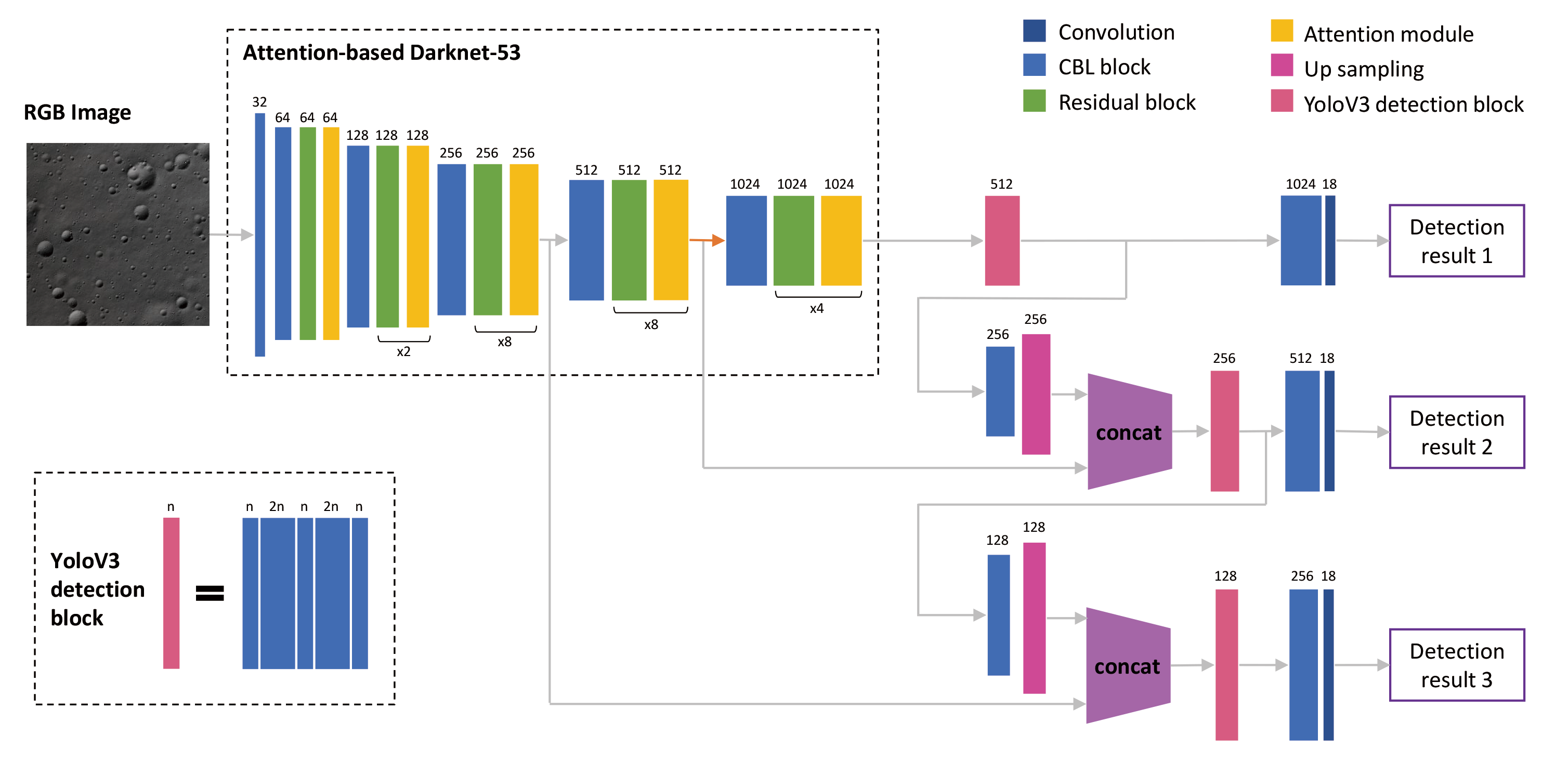}
    \caption{Proposed attention-based Yolov3 framework for Lunar crater detection during landing}
    \label{fig:detection}
\end{figure*}

The YOLOv3 network has a structure consisting of a backbone, a neck, and a head, with a function of extracting features, combining features from different scales, and predicting the location and class, respectively.
For explainable crater detection, we first revise the structure of YOLOv3 and introduce a modified version which replaces the backbone with the proposed attention-based Darknet53.
Therefore, the attention-based YOLOv3 network is offered in Fig. \ref{fig:detection}. The network takes \gls{rgb} images with the size of 256$\times$256 pixels as input and outputs the detected craters with position and radii in the image frame. Furthermore, the loss function of the Lunar crater network includes the bounding box position error $\mathcal{L}_{ciou}$ and the bounding box confidence error (objectness error) $\mathcal{L}_{obj}$.

{\small
\begin{equation}
\begin{split}
    &\mathcal{L}_{obj} =  -\sum_{i=0}^{S^2} \sum_{j=0}^B \mathbbm{1}_{i j}^{\text{obj }}\left[\hat{C}_i \log \left(C_i\right)+\left(1-\hat{C}_i\right) \log \left(1-C_i\right)\right] \\
    &\qquad -\lambda_{\text {noobj }} \sum_{i=0}^{S^2} \sum_{j=0}^B \mathbbm{1}_{i j}^{\text {noobj }} \left[\hat{C}_i \log \left(C_i\right)+\left(1-\hat{C_i}\right) \log \left(1-C_i\right)\right] \\
    &\mathcal{L}_{ciou} =  \sum_{i=0}^{S^2} \sum_{j=0}^B \mathbbm{1}_{i, j}^\text{obj}\left[1-I o U+\frac{\rho^2\left(\bm{b}, \bm{b}^{\text{gt}}\right)}{c^2}+\alpha v\right] \\
    &v =\frac{4}{\pi^2}\left(\arctan \frac{w^{\text{gt}}}{h^{\text{gt}}}-\arctan \frac{w}{h}\right)^2 \\
    &\alpha=\frac{v}{(1-I o U)+v}
\end{split}
\label{eq:loss}
\end{equation}}
where $\mathcal{L}_{ciou}$ is \gls{ciou}, taking into account the overlap area, distance, and aspect ratio. $S^2$ and $B$ represent $S\times S$ grids and prior frame. $\mathbbm{1}_{ij}^{\text{obj}}$ and $\mathbbm{1}_{ij}^{\text{nobj}}$ denote the $j$th bounding box predictor in cell $i$. For $\mathbbm{1}_{ij}^{\text{obj}}$ and $\mathbbm{1}_{ij}^{\text{nobj}}$, if the $j$th predictor in the $i$th cell, set to 1 and 0 respectively, and if there is no predictor, take 0 and 1 respectively. $C_i$ and $\hat{C}_i$ are the confidence levels of the prediction and tagging boxes. $\rho(\cdot)$ is Euclidean distance. The $IoU$ is defined by the predicted ($A$) and groundtruth (B) bounding boxes, given as
\begin{equation}
    IoU = \frac{|A\cap B|}{|A\cup B|}
\end{equation}
Additionally, $\bm{b}, w, h$ and $\bm{b}^{\text{gt}}, w^{\text{gt}}, h^{\text{gt}}$ in Eq. \ref{eq:loss} are the centre coordinates, width, and height of the prediction box and groundtruth box, respectively. $v$ is the consistency of the aspect ratio of \gls{ciou}, and $\alpha$ is a trade-off parameter determined automatically. Since the studied dataset in this paper is circular craters, the prior knowledge of the crater shape is introduced to simplify the $v$, with
\begin{equation}
        v =\frac{4}{\pi^2}\left(\arctan 1-\arctan \frac{w}{h}\right)^2 
\end{equation}

Therefore, the loss function $ \mathcal{L}$ of attention-based YOLOv3 for crater detection is,

\begin{equation}
    \mathcal{L} = \lambda_{iou}\mathcal{L}_{ciou} + \lambda_{obj}\mathcal{L}_{obj}
\end{equation}
where $\lambda_{iou}$ and $\lambda_{obj}$ are the weight factors and are set as 0.05 and 1.0 in this work. 

\subsection{Attention-based RCNN for relative pose estimation}
The features learned by the attention-based \gls{cnn} are post-processed by a deep \gls{rnn} module that models the intrinsic temporal correlations from an ordered sequence of image inputs. This addition is expected to be beneficial to the problem of spacecraft pose estimation due to the inherent relative motion dynamics entailed, and the estimate of the solution for the current frame can benefit from the knowledge of previous frames. Even more than in ground-based applications, the perceived motion of a space target is a smooth function of the previous states instead of changing abruptly.

\begin{figure}[hbt]
    \centering
    \includegraphics[width=0.6\linewidth]{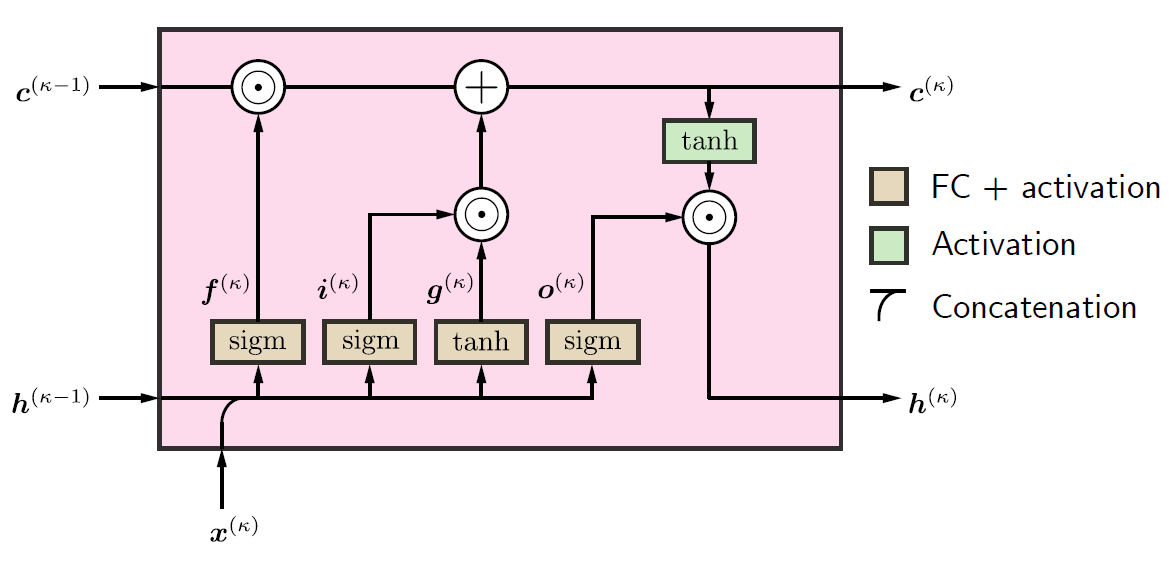}
    \caption{Block diagram of a \gls{lstm} recurrent memory unit.}
    \label{fig:lstm}
\end{figure}

The recurrent feature post-processing module here utilises the \gls{lstm} architecture. The \gls{lstm} combats vital flaws in the capability of vanilla recurrent cells to model long sequences, which are vanishing and exploding gradients. This benefits from the gated design of \gls{lstm} that determines which sectors of the previous hidden state should be kept or discarded in the current iteration. The capability of \gls{lstm} to learn long-term dependencies is achieved not only in combination with the current input, processed by four different units but also by a cell state that acts as an ”information motorway” that bypasses the cells. The \gls{lstm} structure is illustrated in Fig. \ref{fig:lstm}, where $sigm$ and $tanh$ denote the Sigmoid and hyperbolic tangent activation functions. The $\times$ and $+$ denote element-wise product and addition, respectively.
 
\begin{figure*}[htb]
    \centering
    \includegraphics[width=0.6\linewidth]{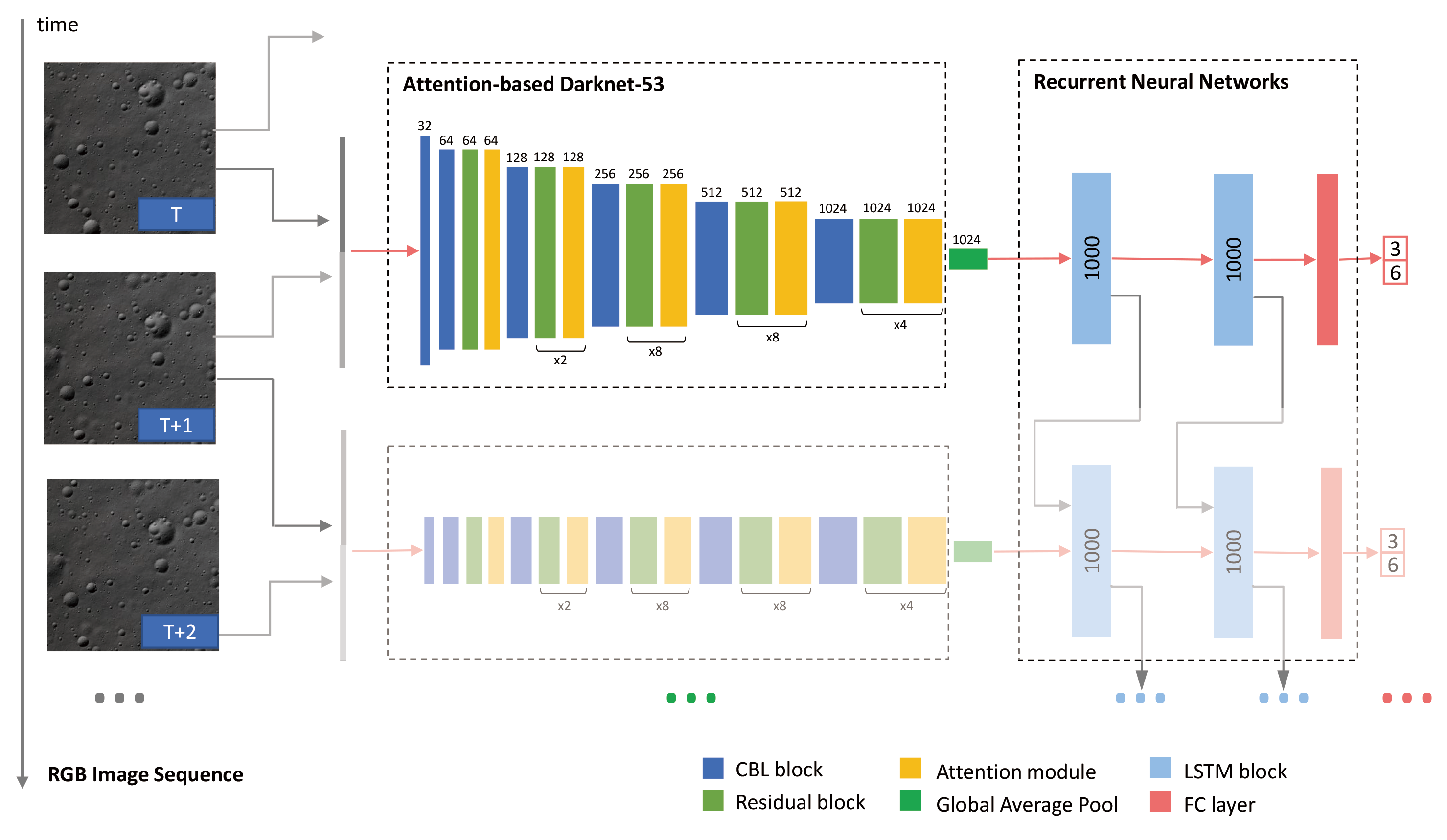}
    \caption{Proposed attention-based \gls{rcnn} framework for relative lunar landing navigation}
    \label{fig:pipeline}
\end{figure*}

Previous studies have shown that two stacked \gls{lstm} layers with 1000 hidden states each are converged and effective in modelling complex motion dynamics. We, therefore, use the same number of layers and hidden neural as DeepVO \cite{wang2017deepvo} and DeepLO \cite{kechagias2020deeplo}. The developed attention-based \gls{rcnn} network consists of an attention-based Darknet53 for feature extraction and a two-layer \gls{lstm} module for sequential modelling with relative pose estimation. Fig. \ref{fig:pipeline} illustrates the overall attention-based \gls{rcnn} structure, and Table \ref{tab:my_label} lists the configuration of the proposed framework. The network takes a monocular \gls{rgb} image sequence as input. At each time step, two consecutive images with the size of $256\times256$ pixels are stacked together to form a tensor in $512\times256\times3$ as the input of the proposed navigation network. Each image pair yields a pose estimate at each time step through the network.

\begin{table}[htb]
\caption{Attention-based navigation network configuration}
    \centering
    \begin{tabular}{cclccc}
    \toprule
    \multicolumn{3}{c}{Layer type} & Filter number & Filter size & Stride \\
    \midrule
    \multicolumn{3}{l}{CBL block} & 64 & 3$\times$3 &2 \\
    \midrule
    \multirow{4}{*}{Residual  block 1} &\multirow{4}{*}{ 1$\times$}     & CBL block & 32 & 1$\times$1 & 1\\
    && CBL block & 64 & 3$\times$3 & 1\\
    && Residual & & & \\
    && Attention module & & & \\
    \midrule
    \multicolumn{3}{l}{CBL block} & 256 & 3$\times$3 &2 \\
    \midrule
    \multirow{4}{*}{Residual block 3 } &\multirow{4}{*}{ $8\times$ } 
    & CBL block & 128 & 1$\times$1 & 1\\
    && CBL block & 256 & 3$\times$3 & 1\\
    && Residual & & & \\
    && Attention module & & & \\
    \midrule
    \multicolumn{3}{l}{CBL block} & 512 & 3$\times$3 &2 \\
    \midrule
    \multirow{4}{*}{Residual  block 4} &\multirow{4}{*}{ $8\times$ } 
    & CBL block & 256 & 1$\times$1 & 1\\
    && CBL block & 512 & 3$\times$3 & 1\\
    && Residual & & & \\
    && Attention module & & & \\
    \midrule
    \multicolumn{3}{l}{CBL block} & 1024 & 3$\times$3 &2 \\
    \midrule
    \multirow{4}{*}{Residual  block 5} &\multirow{4}{*}{ $4\times$ } 
    & CBL block & 512 & 1$\times$1 & 1\\
    && CBL block & 1024 & 3$\times$3 & 1\\
    && Residual & & & \\
    && Attention module & & & \\
    \midrule
    \multicolumn{3}{l}{Global AveragePool} & & & \\
    \multicolumn{3}{l}{\gls{lstm} 1} & \multicolumn{3}{l}{1000 hidden states, sigmoid} \\
    \multicolumn{3}{l}{\gls{lstm} 2} & \multicolumn{3}{l}{1000 hidden states, sigmoid} \\
    \multicolumn{3}{l}{\gls{fl}} & \multicolumn{3}{l}{9$\times$1 output} \\
    \bottomrule
    
    \end{tabular}
    
    \label{tab:my_label}
\end{table} 
\subsection{Training strategies}
\label{sec:training strategies}

\subsubsection{Transfer learning}
Firstly, we utilise the transfer learning technique, in which some network layers are pretrained on a set of tasks involving a large dataset and then used to initialise a same-sized network to solve the purported task. Transfer learning is advantageous for \glspl{cnn} as these typically entail millions of parameters and thus may converge towards a sub-optimal solution if the training data is not diverse enough. 

\begin{figure}
    \centering
    \includegraphics[width=0.6\linewidth]{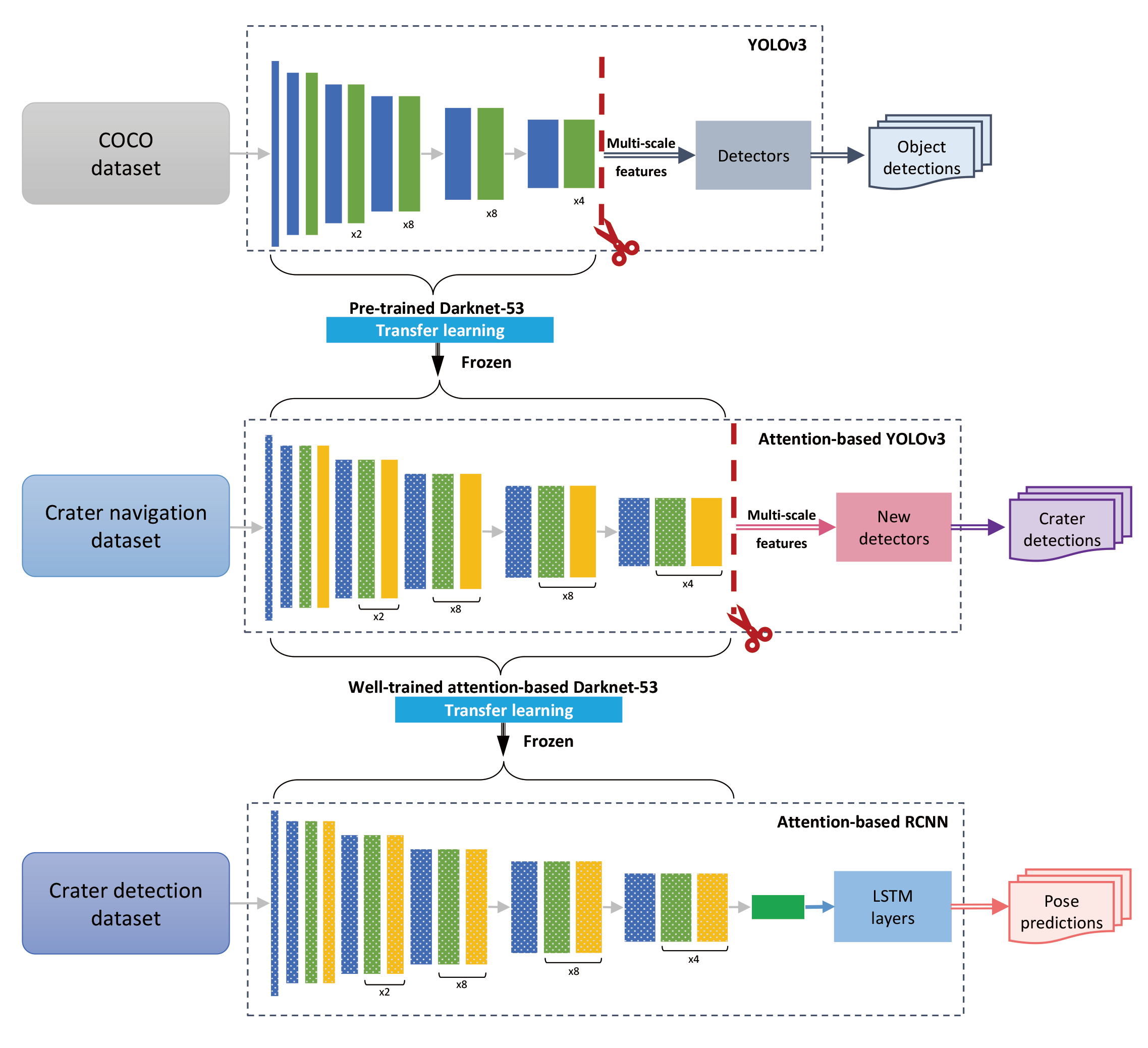}
    \caption{Applied transfer learning in this work.}
    \label{fig:transfer-learning}
\end{figure}

Considering that the data are \gls{rgb} images with three channels as input, we take the pretrained Darknet-53 on \gls{coco} dataset from the second layer to the Global AveragePool layer. Fig. \ref{fig:transfer-learning} shows the applied transfer-learning strategy in this work. 
As shown in Fig. \ref{fig:transfer-learning}, we take the pre-trained Darknet53 layers from YOLOv3 and freeze them in the attention-based YOLOv3 when training for crater detention. On obtaining the well-trained attention-based Darkent-53 layers, they are then extracted and frozen in attention-based \gls{rcnn} for pose predictions.

\subsubsection{Coarse-to-ﬁne training for pose estimation network}
The pose estimation task is relatively more complex than the crater detection task. For fast training convergence, a coarse-to-fine network training approach is applied in training the pose estimation network. 

In the coarse training stage, the layers from well-trained attention-based Darknet53 are frozen and the CNN’s output connects the \gls{fl} and \gls{lstm} layers straightaway. Therefore, the model learns the coarse features using a discredited pose representation. The loss function is formulated as a multi-task learning problem computed by the \gls{mse} of all positions and orientations. Given the poses as $\bm{Y}_t = (\bm{y}_1, \ldots,\bm{y}_t )$, the corresponding sequence of monocular \gls{rgb} images $\bm{X}_t = (\bm{x}_1, \ldots, \bm{x}_t)$, the pose estimation network is to compute the conditional probability of $\bm{Y}_t$ from $\bm{X}_t$ in the probabilistic perspective,
\begin{equation}
    p(\bm{Y}_t|\bm{X}_t) = p(\bm{y}_1, \ldots,\bm{y}_t|\bm{x}_1, \ldots, \bm{x}_t)
\end{equation}
The optimal parameters $\bm{\theta}^*$ for this task is
\begin{equation}
\begin{split}
    \bm{\theta}^* &= \text{argmax}p(\bm{Y}_t|\bm{X}_t;\bm{\theta})  \\
    &= \text{argmin}\frac{1}{N}\sum_{i=0}^N \parallel\hat{\bm{P}}_k -\bm{P}_k\parallel^2_2 + \kappa\parallel\hat{\bm{\varphi}}_k - \bm{\varphi}_k\parallel_2^2
\end{split}
\end{equation}
where $\parallel \cdot \parallel_2$ is a two-norm operator, $\kappa$ is a scale factor to balance the weights of positions and orientations, and $N$ is the number of samples. The orientation $\varphi$ is represented by Euler angles. The loss function of the coarse training stage is then written as
\begin{equation}
    \bm{L}^{(S1)} = \frac{1}{N}\sum_{i=0}^N \sum_{k=0}^t \{\parallel\hat{\bm{P}}_k -\bm{P}_k\parallel^2_2 + \kappa\parallel\hat{\bm{\varphi}}_k - \bm{\varphi}_k\parallel_2^2\}
\end{equation}

To fine-tune the parameters, we unfreeze the frozen layers and train the proposed network of the whole structure. In this ﬁne training stage, the network weights are initialised with the results of coarse training. Instead, the loss function of fine training is redefined based on the Bayesian framework detailed in \cite{zhou2019continuity}, in which the weights $\{ \sigma_p, \sigma_\varphi\}$ in the loss are learnable and iterated with the training process.
\begin{gather}
        \bm{L}^{(S2)} = \bm{L}^{(S2)}_{\bm{p}} e^{-\sigma_p} + \bm{L}^{(S2)}_{\bm{\varphi}}e^{-\sigma_\varphi} + \sigma_p + \sigma_\varphi \\
        \bm{L}^{(S2)}_{\bm{p}} = \sum_{i=0}^N \sum_{k=0}^t \parallel\hat{\bm{P}}_k -\bm{P}_k\parallel^2_2  \\
        \bm{L}^{(S2)}_{\bm{\varphi}} = \sum_{i=0}^N \sum_{k=0}^t \parallel\hat{\bm{\varphi}}_k - \bm{\varphi}_k \parallel_2^2
\end{gather}

\subsection{Evaluation metrics}
For evaluating the performance of the crater detection network, precision, recall, F1 score, and \gls{ap} are used to analyse detected and ground-truth craters. 

Precision ($P$) is the proportion of the positive class predictions that are actually correct, whereas recall ($R$) is the percentage of relevant results that are correctly classified by the model. The F1 score considers both precision ($P$) and recall ($R$) to calculate the accuracy of model predictions. AP summarises \gls{prc} curve as the weighted mean of precision achieved at each threshold. The mathematical expression of $P$, $R$, $F_1$, and \gls{ap} is given as,
\begin{equation}
    \begin{split}
        &P = \frac{T_p}{T_p + F_p} \\
        &R = \frac{T_p}{T_p + F_n} \\
        &F_1 = \frac{2PR}{P+R} \\
        &AP= \sum_{n}(R_n -R_{n-1})P_n
    \end{split}
\end{equation}
where, $R_n$ and $P_n$ are the precision and recall at the $n$th threshold. A pair $(R_n,P_n)$ is referred to as an operating point in \gls{prc} plot. 

Moreover, the \gls{rmse} is used to analyse the performance of pose estimation, which is defined as,
\begin{equation}
    RMSE = \sqrt{\frac{1}{N} \sum_{k=1}^N (y_k - \hat{y}_k)^2}
\end{equation}
where $y_k$ and $\hat{y}_k$ are the predicted and groundtruth variables. $N$ is the total number of the variable. 

The explainability of the proposed networks is achieved by extracting and visualising the attention maps, which show where the attention layers focus and how the focus changes when the layers go deeper. Additionally, to quantitively evaluate the explainability of proposed networks, the \gls{pcc} is introduced in our work, with
\begin{equation}
\rho(a,b)=\frac{\sum_{i=1}^n\left(a_i-\bar{a}\right)\left(b_i-\bar{b}\right)}{\sqrt{\sum_{i=1}^n\left(a_i-\bar{a}\right)^2} \sqrt{\sum_{i=1}^n\left(b_i-\bar{b}\right)^2}}
\end{equation}

where, $\rho$ is the Pearson correlation coefficient, $n$ is sample size, $a_{i},b_{i}$ are the individual sample points indexed with $i$, ${\textstyle {\bar {a}}={\frac {1}{n}}\sum _{i=1}^{n}a_{i}}$ (the sample mean); and analogously for ${\bar {b}}$. 

\section{Landing scenario simulation}
\label{sec:scenario definition}
\begin{figure}[tbh]
    \centering
    \includegraphics[width=0.6\linewidth]{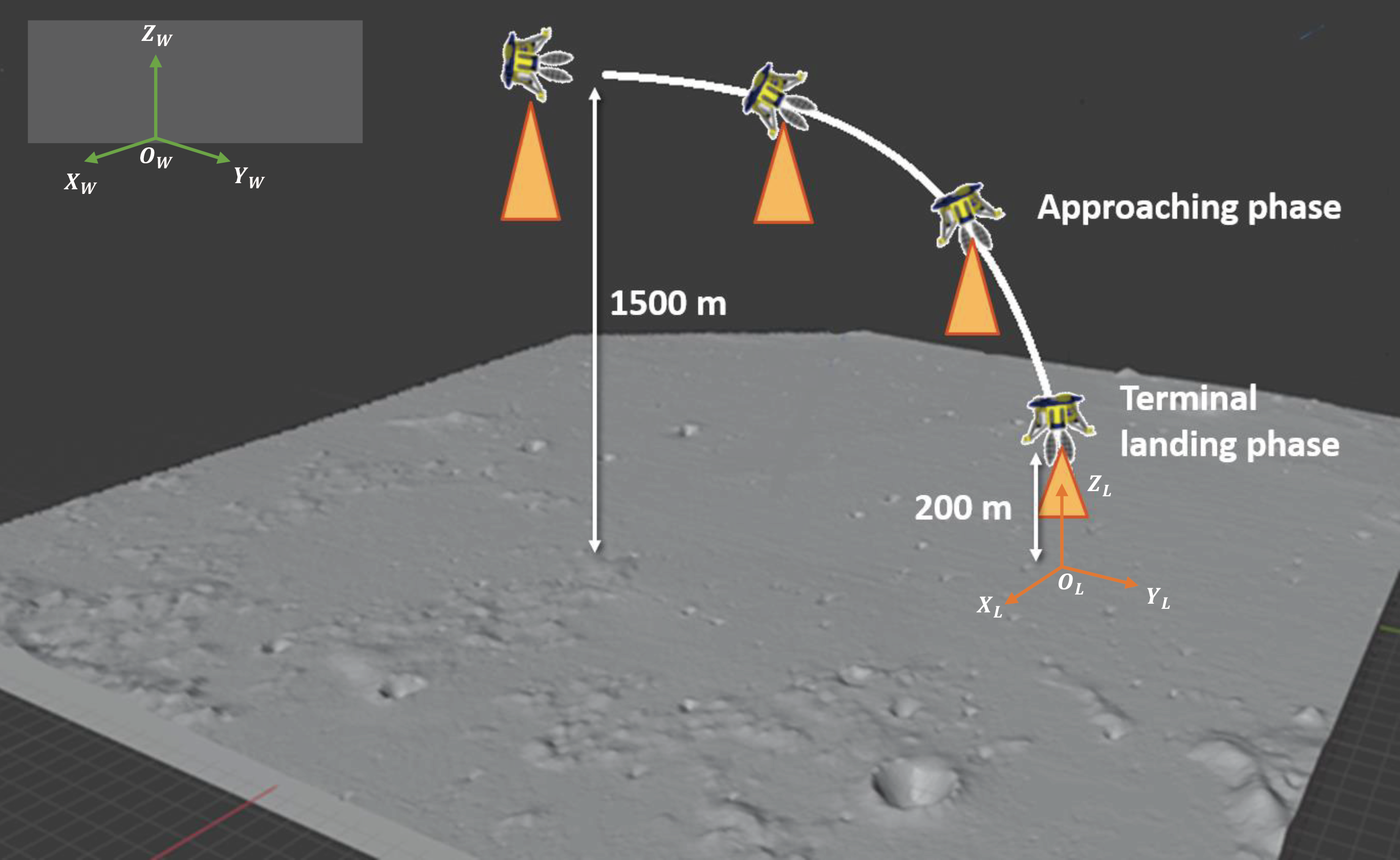}
    \caption{Lunar landing scenario studied in this work}
    \label{fig:scenario}
\end{figure}

The Lunar landing scenario studied in this work is shown in Fig. \ref{fig:scenario} and described as follows: (1) The lander starts at the altitude of \SI{1500}{\meter} with a given coarse absolute pose state. (2) During the landing phase, onboard visible camera keeps working, capturing \gls{rgb} images for navigation. (3) The simulated landing phase is from the altitude of \SI{1500}{\meter} to \SI{200}{\meter}, along a parabola trajectory.

\subsection{Coordinates}
In a lunar landing scenario, five coordinates or frames are introduced, including the \gls{mcmf} frame (specifically the selenographic coordinate system $O_S$), the landing site coordinate $O_L$, the world coordinate $O_W$, the lander body coordinate frame $O_B$, and the lander camera coordinate frame $O_C$. The $O_S$ describes the location and radius parameters of craters on the lunar surface with specified longitude and latitude. The trajectory of the lander is described in $O_L$, in which the origin is the ideal landing location on the surface of the moon, the $Z_L$ axis is perpendicular to the lunar surface in an upward manner, the $Y_L$ axis points to the direction of the southern pole of the moon, and the $X_L$ axis is determined according to the right-hand coordinate system. The landing site coordinate is moon-ﬁxed, and the local landing area studied in this work is seen as a plane, in which all the craters are coplanar. 

\begin{figure}[htb]
    \centering
    \includegraphics[width=0.6\linewidth]{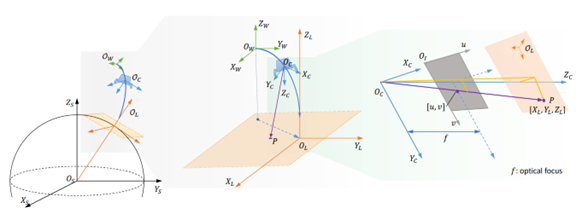}
    \caption{Deﬁnition and relationship of different coordinates for Lunar landing navigation}
    \label{fig:frames}
\end{figure}

Additionally, the $O_W$ is assumed at the initial position of the lander with the same axis direction of the $O_L$ for simplifying the calculation. The $O_B$ is fixed in both origin and orientation to the moving lander and is utilised to provide the pose information of the craft in the given coordinate frame with a given coordinate. The conﬁguration matrix of a camera is the coordinate transformation matrix between $O_C$ and $O_B$ and is commonly known. Fig. \ref{fig:frames} shows the deﬁnition and relationship of the coordinates in the lunar landing navigation scenario. Pose estimation can be realised by computing the transformation matrix of the body coordinate $O_B$ and the landing coordinate $O_L$.

\subsection{PANGU simulation and synthetic data generation}
The synthetic lunar landing scenario is built with PANGU (Planet and Asteroid Natural Scene Simulation Utility) software, where the PANGU is capable of creating models of the planet's surface with given craters and bulks and generating images that are similar to real onboard cameras.

In PANGU, the lunar surface is modelled by the surface modeller function, which accesses the locations and sizes of craters in the \gls{3d} model. The camera properties, position, and orientation can be set by the viewer function. For the trajectories of the lander, a standard parabolic curve is created initially, and then small perturbations with Gaussian distribution are added to it. In total, 49 trajectories are generated, with 100 images captured along each trajectory along the negative axis of $Z_L$ at the given 100 frames. Additionally, a larger dataset of 343 trajectories is rendered for crater detection, with 100 images of each trajectory. All images are in the size of $256\times256$ pixels. 
To label and generate the groundtruth craters in each frame, we calculate the projective point in \gls{2d} image from the known camera internal parameters and camera poses. The craters are then filtered, with radii between 2-50 pixels for crater identification.

\section{Experiments and results}
\label{sec:experiments}
The performance of the proposed attention-based crater detection and lunar landing networks is evaluated on the PANGU synthetic dataset. The proposed network is implemented from the ground up on Pytorch and trained on Quadro GTX 5000 GPU with 16GB RAM. The network is finally tested on Inter Xeon W-10855M CPU with 64-GB RAM to study the running time for further onboard deployment.

\subsection{Performance of crater detection network}
As discussed in \ref{sec:training strategies}, we apply transfer learning to train the proposed attention-based crater detection network by taking advantage of the trained YOLOv3 on the COCO dataset for object detection tasks. K-means clustering is first used to obtain the adaptive anchor sizes from the labelled craters of our landing dataset. Next, we freeze the pre-trained Darknet-53 layers and parameters in the proposed attention-based YOLOv3 and train the rest layers by using our dataset. Finally, we unfreeze the layers and fine-tune the whole proposed crater detection network.

\begin{figure}
    \centering
      \subfloat[\label{1a}]{%
       \includegraphics[width=0.2\linewidth]{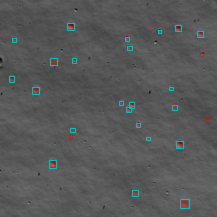}}  \hfill   \subfloat[\label{1b}]{%
        \includegraphics[width=0.2\linewidth]{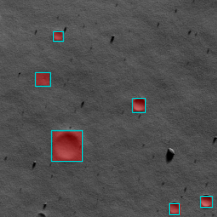}}
  \hfill \subfloat[\label{1c}]{%
        \includegraphics[width=0.2\linewidth]{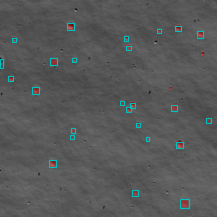}}
  \hfill \subfloat[\label{1d}]{%
        \includegraphics[width=0.2\linewidth]{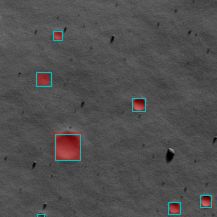}}
  \caption{Crater detection result. (a) and (b) are the results of retrained YOLOv3 for the small and large crater cases. (c) and (d) are the results of the proposed Attention-based YOLOv3 for the small and large crater cases.}
  \label{fig:craterdetection} 
\end{figure}

The PANGU crater detection dataset is split into a $80\%-10\%-10\%$ train-valid-test sub-dataset by randomly shuffling the whole 34300 images. The YOLOv3 is re-trained and evaluated by our dataset as the baseline to compare the performance with the proposed attention-based YOLOv3 for crater detection. Fig. \ref{fig:craterdetection} shows examples of crater detection results in different frames, where the red and cyan rectangles represent the groundtruth and detected craters. It can be seen that small craters are detected slightly better in terms of number and size by introducing the attention mechanism in YOLOv3, whereas the prediction size of large craters seems slightly smaller than the groundtruth. Table \ref{tab:craterdetect} lists the analytical performance of the crater detection network, suggesting that the attention-YOLOv3 outperforms the YOLOv3 in terms of detection precision, recall, AP, and F1 score.

\begin{table}[htb]
    \centering
    \caption{Performance of crater detection network}
    \begin{tabular}{cccccc}
        \toprule
        Network &  precision & recall & AP & F1 score \\
    \midrule
        YOLOV3 & 0.8395 & 0.9258 & 0.8926 & 0.8805\\
        Attention-YOLOV3 & 0.8458 & 0.9336 & 0.8963 & 0.8875 \\
        \bottomrule
    \end{tabular}

    \label{tab:craterdetect}
\end{table}

\subsection{Performance of navigation network}
Similarly, a train-valid-test split is performed on the navigation dataset according to $80\%-10\%-10\%$ split proportion.
Thus, 39 trajectories are used to train attention-based \gls{rcnn} for landing navigation. 
The pipeline is trained on Quadro GTX 5000 GPU with 16GB RAM, in which the sampled length of frames and batch size are set to be 16 and 8, respectively. The learning rate is $1\times10^{-4}$. For the ﬁne training stage, the pipeline is trained with a mini-batch size of 8 and epochs of 20. A cyclical learning rate decay of 5 cycles is adopted for training, with $1\times10^{-4}$ as the initial state. 

A baseline model (Darknet53-\gls{lstm}) is also trained with the same hyperparameters for comparison. At each time step, two consecutive images are stacked together to form a tensor as the input, and each image pair yields a pose estimation through the network. The network input is a stacked \gls{rgb} image with $512\times256\times3$. The network output is a $9\times1$ vector, with $3\times1$ representing the relative position and $6\times1$ representing relative rotation. The rotation is described by the continuity rotation representation detailed in \cite{zhou2019continuity}.

\begin{figure}[htb]
    \centering
    \includegraphics[width=0.6\linewidth]{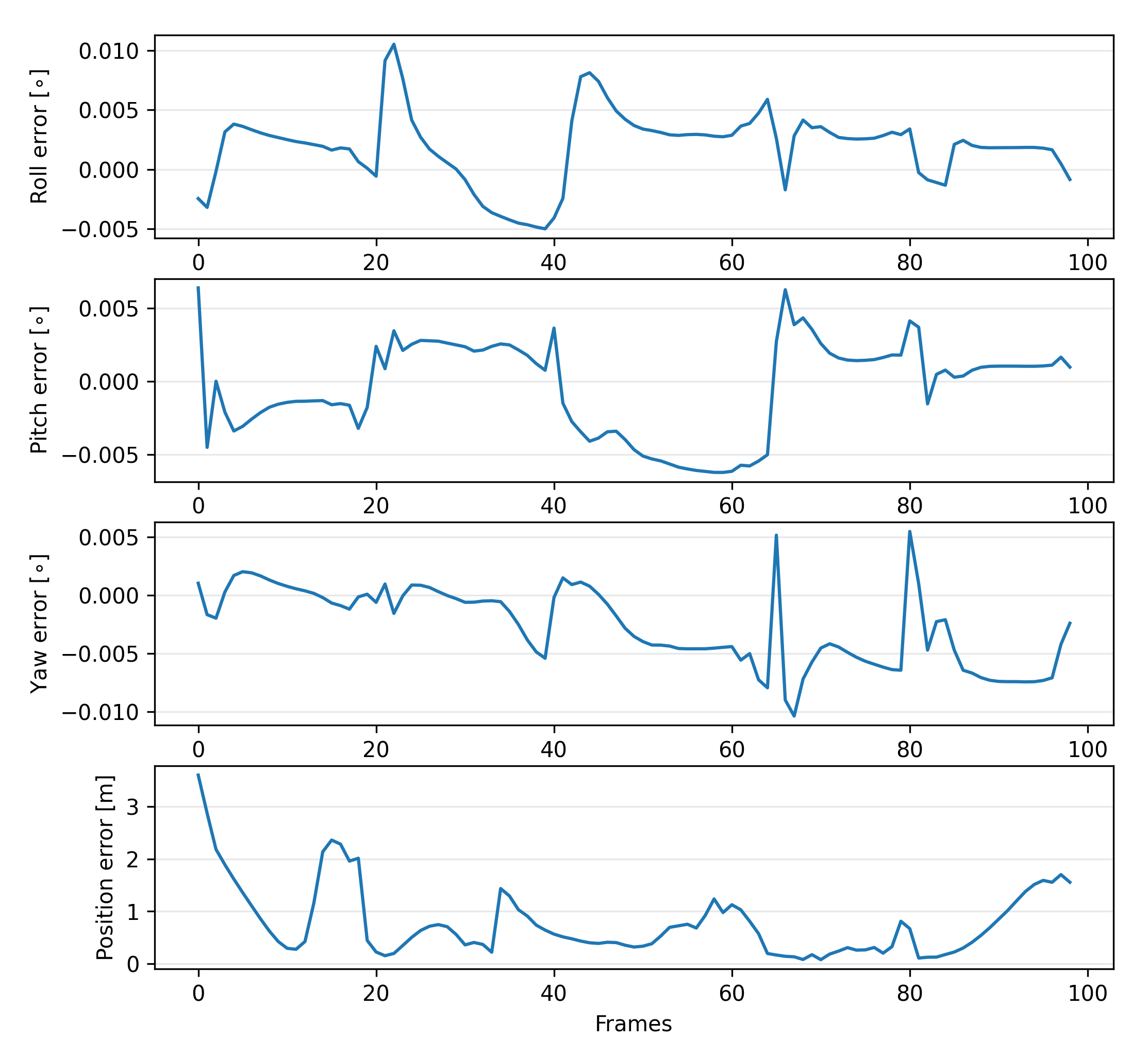}
    \caption{Pose estimation of lunar landing scenario in PANGU}
    \label{fig:frames_err}
\end{figure}

Table \ref{tab:navigationresult} shows the comparison between the baseline and proposed navigation network. The attention-based Darknet53-\gls{lstm} network performs better than the baseline model with the position \gls{rmse} of $\sim$ \SI{1.14}{\meter}. The running time for pose estimation from each input is $\sim$\SI{0.22}{s} same as the baseline model, which indicates the capability of performing onboard real-time navigation. Fig. \ref{fig:frames_err} illustrates the pose estimation error of the proposed navigation network on a landing trajectory, showing that the network converges after 20 training epochs. However, some local peaks can be seen in Fig. \ref{fig:frames_err}. This result might be explained by the fact that the testing trajectory frames are larger than the training trajectory frames, with 99 frames compared to 16 frames.



\begin{table}[htb]
    \centering
    \caption{Performance of lunar landing navigation network}
    \begin{tabular}{m{1.5cm}<{\centering}m{1cm}<{\centering}m{1.3cm}<{\centering}m{1.3cm}<{\centering}m{1.5cm}<{\centering}}
    \toprule
        Network &  Training epochs & Running time [fps] & Position \gls{rmse} [m] & Orientation \gls{rmse} [deg] \\
    \midrule
        Baseline: Darknet53-LSTM& 20 & $\sim$0.22 & $\sim$7.61 & $\sim$1.1e-2\\
        Attention-Darknet53-LSTM & 20 & $\sim$0.22 & $\sim$1.14 & $\sim$5.5e-3 \\
        \bottomrule
    \end{tabular}
    \label{tab:navigationresult}
\end{table}

\subsection{Explainability analysis}

\begin{figure}[htb]
    \centering
    \includegraphics[width=0.7\linewidth]{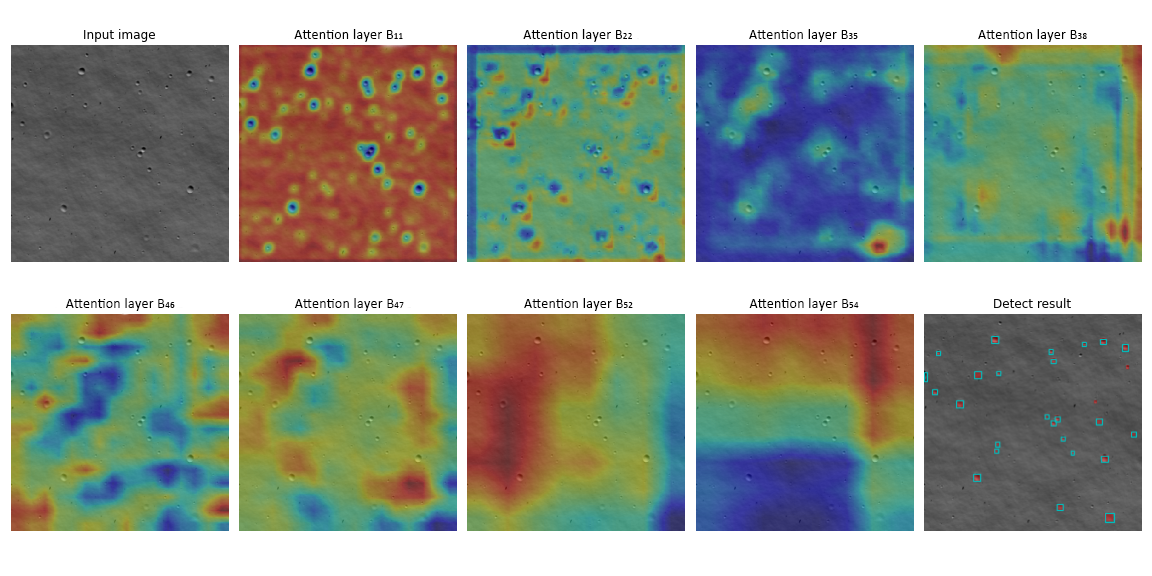}
    \includegraphics[width=0.7\linewidth]{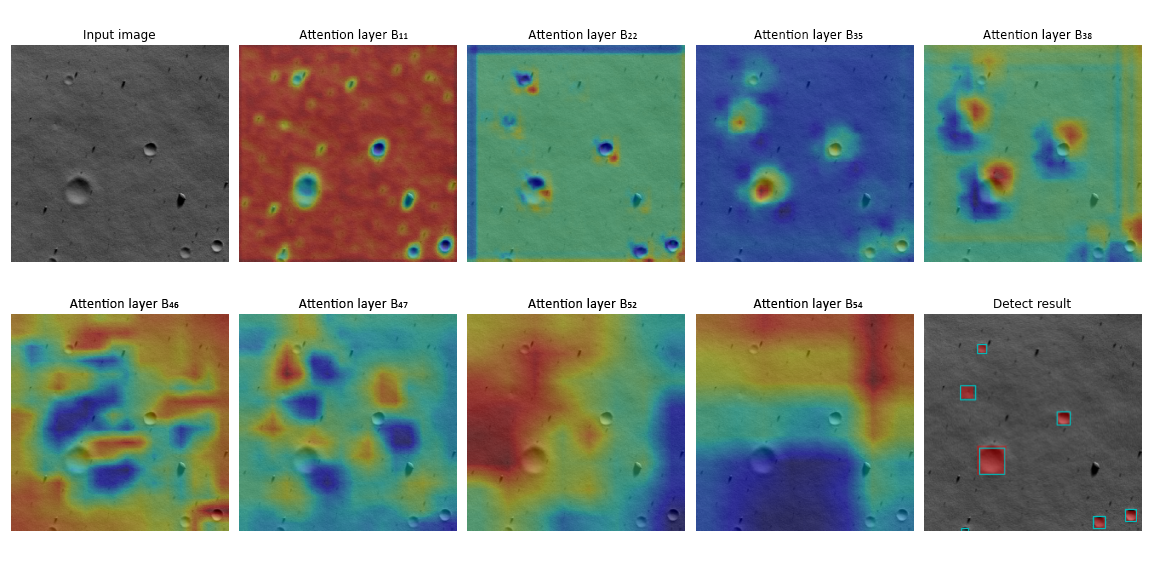}
    \caption{Attention masks at different convolutional layers for crater detection network}
    \label{fig:frames_att}
\end{figure}

The explainability is first achieved by visualising the attention masks in the proposed networks. Fig. \ref{fig:frames_att} shows the attention masks of different convolutional layers for the crater detection network, in which the attention layer name $\text{B}_\text{ij}$ represents the $j$th attention layer in the $i$th Residual block. 

As shown in Fig. \ref{fig:frames_att}, the first attention layer $\text{B}_\text{11}$ indicates that the feature extraction network focuses on flat areas, with the rough areas (craters and boulders) having low values. After completing the first two Residual blocks, the network pays more attention to larger craters and boulders depicted in $\text{B}_\text{22}$. $\text{B}_\text{35}$ and $\text{B}_\text{38}$ show craters in the image frame are highlighted through the first three Residual blocks, whereas $\text{B}_\text{46}$ and $\text{B}_\text{47}$ indicate that there is not a strong relationship between the attention mask and input image. With the feature extraction going deep, the deep attention layers $\text{B}_\text{52}$ and $\text{B}_\text{54}$ seem not to correlate with the input images but have a strong correlation between each with different inputs. A possible explanation for these findings is that different convolutional layers of the proposed network have various functions. The shallow convolutional layers focus on both craters and boulders, then identify craters and boulders, and the deeper layers process the extracted features into output format. Thereby, we have some knowledge of how the proposed network works in crater detection by visualising the attention layers.

To further discuss the findings and conjectures, we quantitatively analyse the correlation between attention masks and groundtruth masks using the \gls{pcc}. Each groundtruth mask is described by a greyscale image using groundtruth crater labels in the image frame, with the background and craters set to be black(``1'') and white(``0''). Additionally, the width of crater edges varies, and the pixel values are defined as,

\begin{equation}
g(x,y)=\left\{
	\begin{aligned}
	0.2 \quad \quad r_k< \parallel (x,y)-(x_k,y_k)   \parallel \leq \alpha_1 r_k  \\
	0.4 \quad  \alpha_1 r_k<\parallel (x,y)-(x_k,y_k)   \parallel \leq \alpha_2 r_k  \\
	0.6 \quad \alpha_2 r_k <\parallel (x,y)-(x_k,y_k)   \parallel \leq \alpha_3 r_k  \\
        0.8 \quad \alpha_3 r_k<\parallel (x,y)-(x_k,y_k)   \parallel \leq \alpha_4 r_k  \\
	\end{aligned}
	\right.
	\label{eq:crater_width}
\end{equation}

where, $(x_k,y_k)$ is the $k$th groundtruth crater centre in image frame and $r_k$ is the corresponding radius. $\{\alpha_1, \alpha_2, \alpha_3, \alpha_4 \}$ are empirical factor given as $\{1.1, 1.2, 1.3, 1.4 \}$ in our paper. All attention masks are resized to the same size as the groundtruth mask with 256$\times$256 pixels. Giving the input image $I^n$ and groundtruth mask $G^n$, the corresponding attention masks $\text{B}_{\text{ij}}^n$ can be obtained. We then compute the \gls{pcc} in two manners, defining as,
\begin{equation}
  \text{PCC}_1(\text{B}_{\text{ij}}) = \frac{1}{N} \sum_{n=1}^{N}\rho(\text{B}_{\text{ij}}^n,G^n) 
\end{equation}
\begin{equation}
  \text{PCC}_2(\text{B}_{\text{ij}}) = \frac{1}{N} \sum_{n=1}^{N}\rho(\text{B}_{\text{ij}}^n,\text{B}_{\text{ij}}^m) \quad m \neq n
\end{equation}

\begin{figure}[htb]
    \centering
    \includegraphics[width=0.6\linewidth]{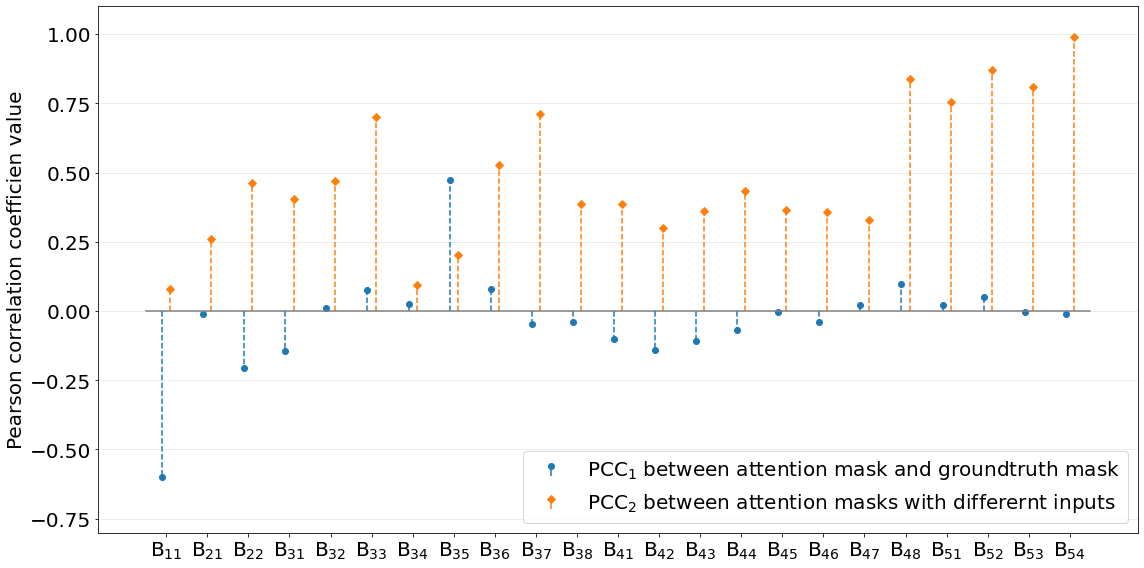}
  \caption{\gls{pcc} for proposed crater detection network. }
    \label{fig:att_pearson_crater}
\end{figure}

The results of the correlational analysis are presented in Fig. \ref{fig:att_pearson_crater}, comparing the $\text{PCC}_1(\text{B}_{\text{ij}})$ and $\text{PCC}_2(\text{B}_{\text{ij}})$ of crater detection network. It can be seen from the data in Fig. \ref{fig:att_pearson_crater} that the shallow convolutional layers $\text{B}_\text{11}$ and $\text{B}_\text{35}$ have a moderate correlation with the groundtruth mask, indicating that these layers focus on craters. The deeper convolutional layers have no correlation with the ground mask but a strong correlation between each other with $\text{PCC}_2(\text{B}_{\text{48}})$,$\text{PCC}_2(\text{B}_{\text{51}})$,$\text{PCC}_2(\text{B}_{\text{52}})$, $\text{PCC}_2(\text{B}_{\text{53}})$, $\text{PCC}_2(\text{B}_{\text{54}}) >0.75$. The values of $\text{PCC}_1(\text{B}_{\text{ij}})$ and $\text{PCC}_2(\text{B}_{\text{ij}})$ suggest the functions of different convolutional layers, and further support the hypothesis that the shallow convolutional layers focus and identify craters, and the deeper layers process the extracted features into output format. Quantitative analysis can help to tell the specific function of different convolutional layers and may further help to modify network structure.

For the explainability of the relative navigation network, we first visualise the attention masks offered in Fig. \ref{fig:frames}. As presented in Fig. \ref{fig:frames}, attention masks from the first two Residual blocks ($\text{B}_\text{11} \sim \text{B}_\text{22}$ and $\text{B}_\text{31}$) show the module focus on craters in the image. Attention masks from the third residual block highlight larger craters and patterns. Subsequently, attention masks from the fourth, fifth, and sixth Residual blocks change attention from centre patterns to horizontal and vertical edges. Moreover, the deeper attention masks can still look at craters to generate features if the crater in the image is large.
\begin{figure}[htb]
    \centering
    \includegraphics[width=0.7\linewidth]{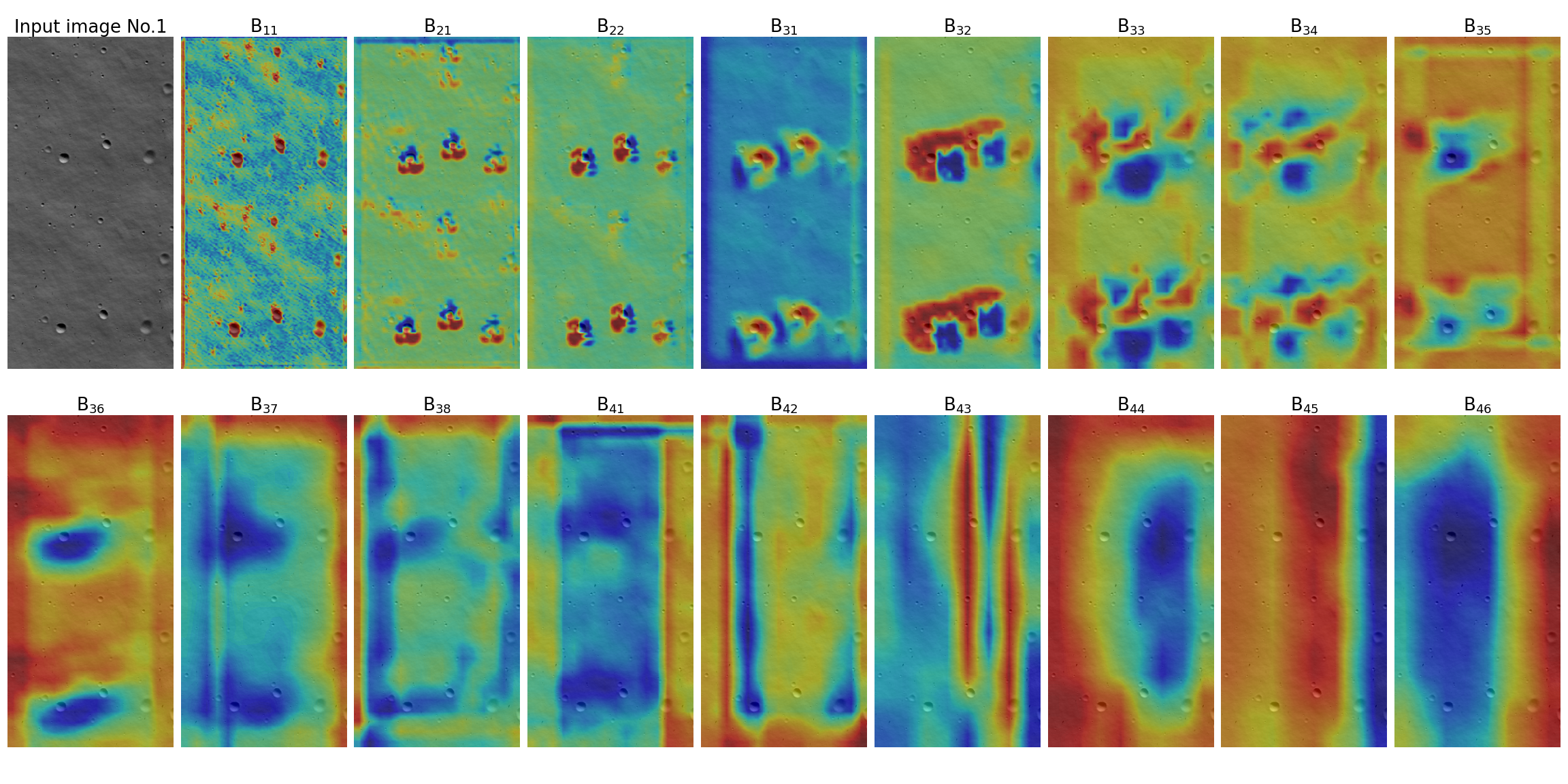}
    \includegraphics[width=0.7\linewidth]{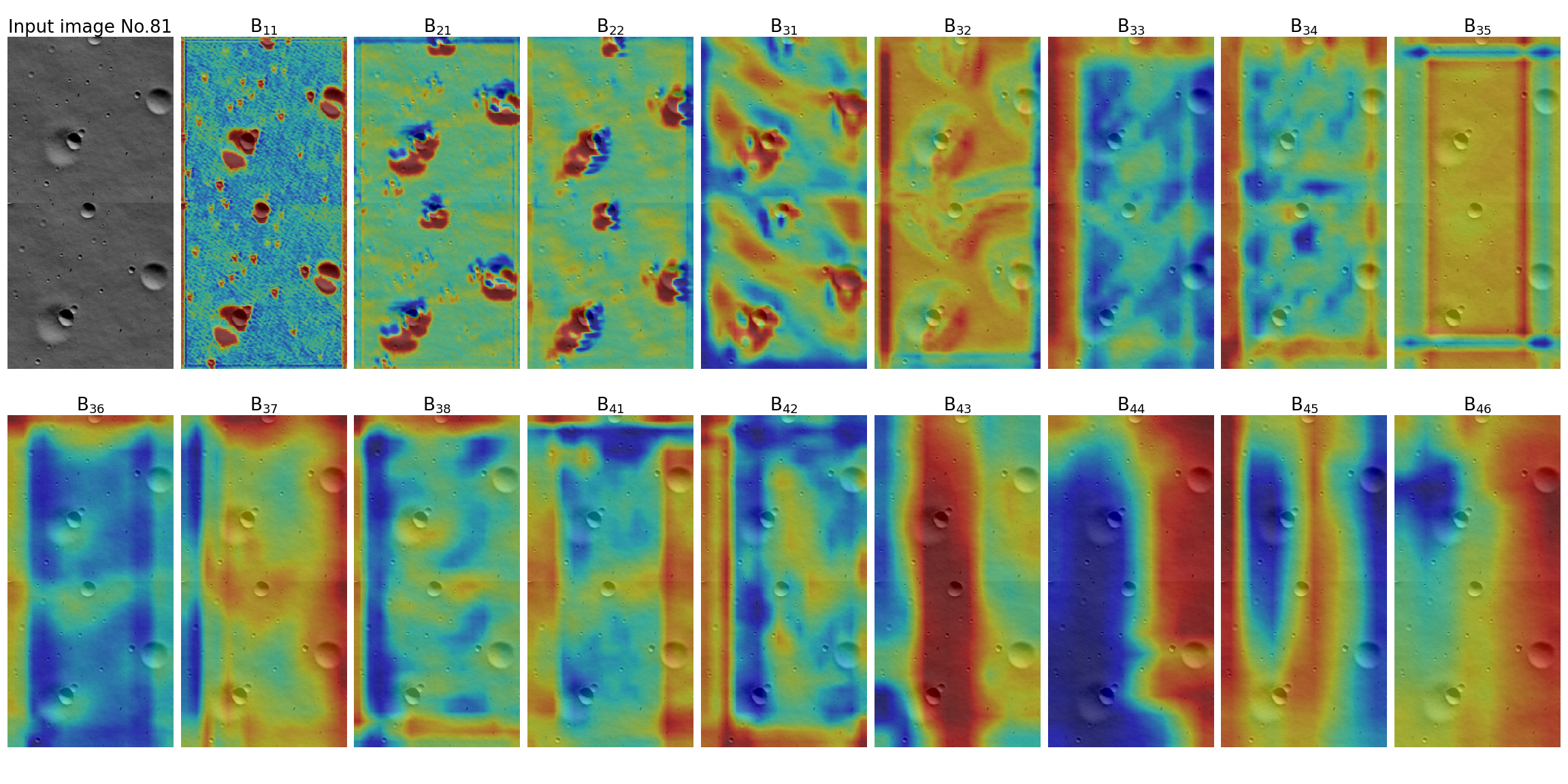}
    \caption{Attention masks of proposed navigation network}
    \label{fig:attmasks}
\end{figure}

\begin{figure}[htb]
    \centering
    \includegraphics[width=0.7\linewidth]{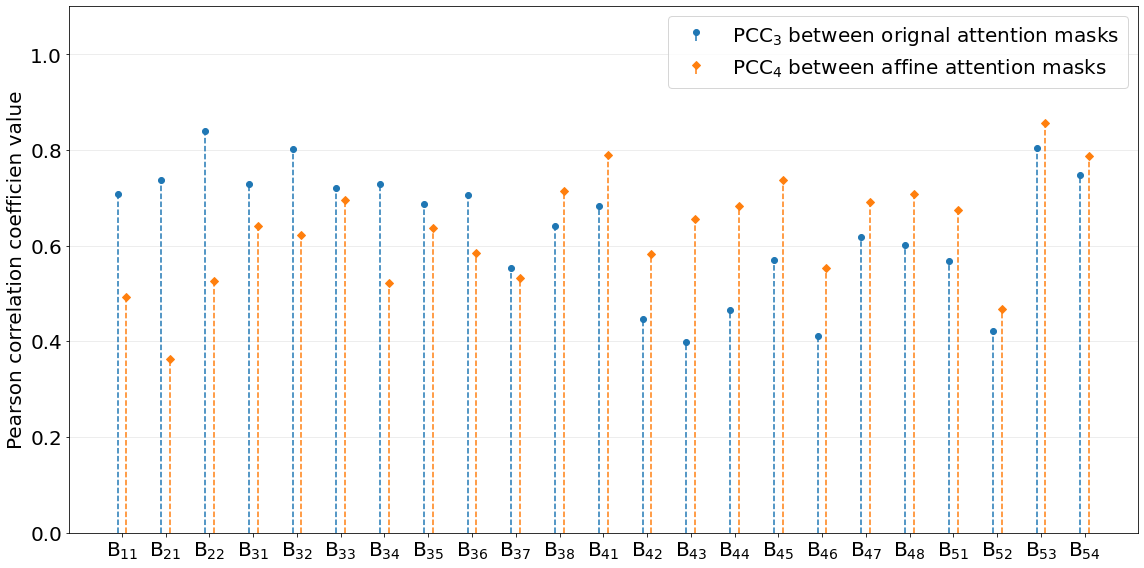}
    \caption{\gls{pcc} for proposed navigation network}
    \label{fig: navigation analysis}
\end{figure}

Similarly, we use the \gls{pcc} to quantitate the correlation between the groundtruth pose and attention masks. As attention masks are images while the groundtruth pose is a vector, we cannot compute the \gls{pcc} as the way crater detection network does directly. Thus, the affine transformation is introduced in calculating the \gls{pcc} for the proposed navigation network. The idea is that, 1) dividing the one attention mask $\text{B}_\text{ij}$ into two images $\text{B}_\text{ij}[u]$ and $\text{B}_\text{ij}[d]$ with the size of 256$\times$256 pixels, 2) applying the affine transformation to $\text{B}_\text{ij}[d]$ and then having the image $\text{B}'_\text{ij}[d]$, 3) computing the \gls{pcc} between the $\text{B}_\text{ij}[u]$, $\text{B}_\text{ij}[d]$, and $\text{B}'_\text{ij}[d]$ with the formula given by,

\begin{equation}
  \text{PCC}_3(\text{B}_{\text{ij}}) = \frac{1}{N} \sum_{n=1}^{N}\rho(\text{B}_{\text{ij}}^{n}[u],\text{B}_{\text{ij}}^{n} [d]) 
\end{equation}
\begin{equation}
    \begin{split}
   \text{PCC}_4(\text{B}_{\text{ij}}) &= \frac{1}{N} \sum_{n=1}^{N}\rho(\text{B}_{\text{ij}}^{n} [u],\text{B}_{\text{ij}}^{'n} [d])  \\
   \text{B}_{\text{ij}}^{'n} [d] &= \bm {M}(\hat{\bm{y}}_n)  \text{B}_{\text{ij}}^{n} [d]
    \end{split}
\end{equation}
where $ \bm {M}(\hat{\bm{y}}_n)$ is the affine transformation matrix defined by groundtruth pose $\hat{\bm{y}}_n$. Thus, $\text{PCC}_3(\text{B}_{\text{ij}})$ and $\text{PCC}_4(\text{B}_{\text{ij}})$ are the \gls{pcc} between the original attention layers and affine attention masks respectively, comparing in Fig. \ref{fig: navigation analysis}. 

Fig. \ref{fig: navigation analysis} reveals that there has been a marked increase in the value of $\text{PCC}_4(\text{B}_{\text{11}}) \sim \text{PCC}_4(\text{B}_{\text{32}})$, $\text{PCC}_4(\text{B}_{\text{34}})$ compared to the corresponding $\text{PCC}_3(\text{B}_{\text{ij}})$ after introducing affine transformation. A strong correlation is found between the affine attention masks in terms of the $\text{PCC}_4(\text{B}_{\text{22}})$ and $ \text{PCC}_4(\text{B}_{\text{32}})$. Comparing the trend of the $\text{PCC}_4$ and $\text{PCC}_3$, it can be seen that the $\text{PCC}_3$ is larger than the $\text{PCC}_4$ after the attention mask $\text{B}_{\text{37}}$, which indicates that a stronger correlation between the original attention masks. Moreover, it suggests the function of different layers of the navigation network, with the shallow layers focusing and extracting features of pose information and the deep layers processing the features into the output format. 

Furthermore, comparing the trend of $\text{PCC}_1$, $\text{PCC}_2$ and $\text{PCC}_3$, $\text{PCC}_4$ in Fig. \ref{fig:att_pearson_crater} and Fig. \ref{fig: navigation analysis}, both networks focus and extract features in the first two Residual blocks and process the features in the fourth and fifth Residual blocks. Additionally, there is an impressive finding that several convolutional layers of the third Residual block still focus and extract features with a strong correlation between the groundtruth, and the number of convolutional layers varies in dealing with different tasks. For complex tasks, such as pose estimation for navigation, more convolutional layers are worked on feature extraction. This is suggested by comparison of $\text{PCC}_4(\text{B}_{\text{36}}) > \text{PCC}_3(\text{B}_{\text{36}})$ and $\text{PCC}_1(\text{B}_{\text{36}}) < \text{PCC}_2(\text{B}_{\text{36}})$. Together these findings provide important insights into understanding how attention mechanism provides explainability to the network for different tasks.  

\section{Conclusion}
\label{sec:conclusion}
This study set out to study the explainable network for vision-based lunar landing. Attention-based Darknet53 is proposed as the feature extraction structure for crater detection and navigation tasks. Attention-based YOLOv3 is presented for crater detection with explainability, and attention-based \gls{rcnn} (attention-Darknet53-LSTM) is offered for relative pose estimation with the input of \gls{rgb} images. The experimental simulation results demonstrate the efficiency and performance of the proposed crater detection and relative navigation solution, and the explainability of the provided networks is achieved by introducing an attention mechanism into the network during model building. By visualising the attention masks at different network layers, we know how the network pays visual attention to various regions of input images. Quantitative analysis between the attention masks, the inputs, and groundtruth is performed by taking advantage of the \gls{pcc}. The quantitative analysis shows the functions of different layers in a network, with the shallow layers focusing and extracting features and the deep layers processing the features into the output format. When using the same backbone structure dealing with different tasks, more convolutional layers are worked on feature extraction if the task is more complex. Taken together, the findings of this research provide insights for the attention mechanism in \gls{xai}. Further research	might explore optimising the \gls{cnn} structure with the attention mechanism.


\section*{Acknowledgement}

This work is supported by the ESA, with Contract No. 4000133292/20/NL/CRS/kk.



\end{document}